\documentclass[preprint,11pt]{elsarticle}
\usepackage[margin=1in]{geometry} 

\usepackage{graphicx}%
\usepackage{multirow}%
\usepackage{amsmath,amssymb,amsfonts}%
\usepackage{amsthm}%
\usepackage{mathrsfs}%
\usepackage[title]{appendix}%
\usepackage{xcolor}%
\usepackage{textcomp}%
\usepackage{manyfoot}%
\usepackage{booktabs}%
\usepackage{algorithm}%
\usepackage{algorithmicx}%
\usepackage{algpseudocode}%
\usepackage{listings}%
\usepackage{array}
\usepackage{subcaption}
\usepackage{booktabs}

\begin{document}

\begin{frontmatter}



\title{Ethio-Fake: Cutting-Edge Approaches to Combat Fake News in Under-Resourced Languages Using Explainable AI}





\author[a]{Mesay Gemeda yigezu $^\ast$} 
\author[b]{Melkamu Abay Mersha}
\author[a]{Girma Yohannis Bade}
\author[b]{Jugal Kalita}
\author[a]{Olga Kolesnikova}
\author[a]{Alexander Gelbukh}

\address[a]{Instituto Politécnico Nacional (IPN), Centro de Investigación en Computación (CIC), Mexico city, Mexico}
\address[b]{College of Engineering and Applied Science, University of Colorado Colorado Springs (UCCS), Colorado Springs, USA}

\begin{abstract}
The proliferation of fake news has emerged as a significant threat to the integrity of information dissemination, particularly on social media platforms. Misinformation can spread quickly due to the ease of creating and disseminating content, affecting public opinion and sociopolitical events. Identifying false information is therefore essential to reducing its negative consequences and maintaining the reliability of online news sources. Traditional approaches to fake news detection often rely solely on content-based features, overlooking the crucial role of social context in shaping the perception and propagation of news articles. In this paper, we propose a comprehensive approach that integrates social context-based features with news content features to enhance the accuracy of fake news detection in under-resourced languages. We perform several experiments utilizing a variety of methodologies, including traditional machine learning, neural networks, ensemble learning, and transfer learning. Assessment of the outcomes of the experiments shows that the ensemble learning approach has the highest accuracy, achieving a 0.99 F1 score. Additionally, when compared with monolingual models, the fine-tuned model with the target language outperformed others, achieving a 0.94 F1 score. We analyze the functioning of the models, considering the important features that contribute to model performance, using explainable AI techniques.
\end{abstract}

\begin{keyword}
Fake News \sep Misinformation \sep Amharic \sep Under-resourced Languages \sep Low-resourced Languages \sep Ethiopian Languages \sep Basic Neural Network \sep Traditional Machine Learning \sep Ensemble \sep Transfer Learning \sep LIME \sep Explainable AI.
\end{keyword}

\end{frontmatter}


\section{Introduction}
\label{intro}

The rise of social media platforms has increased the volume of inter-personal group-based as well as broadcast communication, casual sharing of information, and the overt urge and need to post frequent news and status updates. Facebook, Twitter, YouTube, and Reddit are some of the most widely used social networking platforms to deliver 
information globally \cite{yigezu2023evaluating}. In addition to providing for social interactions, they can enhance the quality of public relations, increase the sense of participation and allow academia, governments, and industry to showcase achievements effectively and efficiently \cite{eyrich2008pr}. However, in many situations, social media content is misleading and attempts to deceive users. Any incorrect or deceptive information that purports to be newsworthy is referred to as fake news \cite{malliga2023overview}.

The spread of unverified false information can have serious repercussions, such as harming the credibility of the news ecosystem, ruining the reputation of any individual or group, or inciting public panic that can undermine societal stability \cite{shu2017fake, yee2017post}. Its downside is that it has the potential to spread quickly and affect people's perceptions of various subjects. It intentionally misleads the audience and potentially causes protests that result in economic, social, and political crises, diverts the attention of decision-makers, and instills fear and bewilderment in others, and it is arguably a significant reason for conflict and war in many countries \cite{babacan2022information, worku2022amharic}.  Although there has been a growth in multilingual web content, fake news classification in low-resource languages is still a challenge in developing countries due to the non-availability of annotated corpora and tools\cite{yigezu2021multilingual}. According to the European Institute of Peace Assessment, fake news, misinformation, and hate speech have flourished in Ethiopia's media ecosystem, especially online. This is significantly linked to serious, tragic events and has contributed to violence and war \footnote{https://www.eip.org/wp-content/uploads/2021/04/Fake-News-Misinformation-and-Hate-Speech-in-Ethiopia.pdf}.

Nowadays, many researchers are paying a lot of attention to combating the issue of fake news. In order to detect fake news, most researchers have focused only on news content. However, detecting fake news on social media poses a considerable challenge, particularly when relying solely on textual features. This difficulty stems from the multifaceted nature of fake news on social media, encompassing various features such as news content, social networking, user-based interactions, and action-based cues, all of which collectively influence the authenticity of the news being disseminated \cite{shu2017fake}. Relying solely on linguistic features is inadequate for detecting fake news \cite{ruchansky2017csi, feng2012syntactic,potthast2017stylometric}. 

Indeed, the features of fake news are linguistically and technically difficult to comprehend, interpret, extract, and analyze \cite{ruchansky2017csi,vosoughi2018spread}. To mitigate these detrimental effects, the development of techniques for automatically detecting fake news on social media is crucial \cite{gereme2021combating,pennycook2019fighting}. Therefore, integrating appropriate news and social context features is essential to enhance the accuracy and speed of fake news detection mechanisms. Using a variety of factors, such as textual content, user engagement patterns, and network dynamics, the detection algorithms can better tell the difference between real and fake information. This makes news platforms more reliable and protects the integrity of public discourse.

This paper discusses the development of fake news detection in a low-resourced Ethiopian language, specifically Amharic, to improve performance for future needs. It is also important to have a reliable fake news detection (FND) method that can mine and use the extra information that comes from social engagement in the network, especially news consumption on online communication platforms. This extra information would help make fake news predictions more accurate. In addition, we evaluate the functionality of the models, including their ability to identify the most influential words in identifying input text as either fake or real, using explainable AI (XAI) methodologies.
As a result, the following points can be taken as contributions:
\begin{itemize}
\item We created a data set of fake news in Amharic from various domains, providing it as a valuable resource for future research.
\item We experiment with FND methods using hybrid features, establishing a set of features that are effective in classifying fake news in Amharic.
\item We compare and analyze different approaches and find that a fine-tuned model with the target language is most effective.
\item We implement XAI to enhance the trustworthiness and transparency of the model we use.
\end{itemize}

\textbf{Defining Fake News problem}

In this section, we explain how we automatically detect fake news on social media. First, we define the important parts of fake news. Then, we give a formal definition of a method to detect fake news. We use some basic notation, as explained below.

Let \( a \) refer to a news article. It consists of two major components: social context and news content. The social context mainly includes Publisher. Publisher \( \overrightarrow{Pa} \) comprises a set of profile features to describe the original author, such as user name, sex, and age, among other attributes. Content \( \overrightarrow{Ca} \) consists of a set of attributes that represent the news and includes headline, text, etc.

We define Social News Engagements as a set of tuples \( E =\{e_{it}\}\) to illustrate how a news article spreads over time among 
\( n \) users \( U = \{u_1, u_2, \ldots, u_n\} \) and their corresponding posts on social media \( P = \{p_1, p_2, \ldots, p_n\} \) regarding the news article \( a \). Each engagement \( e_{it} = \{u_i, p_i, t\} \) indicates that user \( u_i \) shares the news article \( a \) using post \( p_i \) at time \( t \).  If the news article  \( a \) hasn't received any engagement yet, we set \( t = \text{Null} \) and \( u_i \) to represent the publisher.

In our scenario, we have social news engagements \( E \) among \( n \) users for a particular news article   \( a \). The goal of fake news detection is to determine whether this news article \( a \) is fake or not. This is represented as a prediction function \( f \), which maps the set of social news engagements \( E \) to the binary set \{0, 1\}, indicating whether the news article is classified as fake (1) or not fake (0).

\[
f(a) = 
\begin{cases} 
    1, & \text{if } a \text{ contains fake news content} \\ 
    0, & \text{otherwise}
\end{cases}
\]

The rest of the paper is structured as follows: In Section \ref{related}, we review related work on fake news detection, covering various approaches. Section \ref{data} elaborates on our proposed architecture, including a detailed discussion of our dataset. Section 
 \ref{model} explains what models we used. Section \ref{result} provides a detailed analysis of the experimental results. In Section \ref{xai}, we discuss the concept of interpretability modeling, which is employed to bolster the trustworthiness and transparency of the models we utilize. Finally, in Section \ref{conclusion}, we present the conclusions of our research.

\section{Related Work}
\label{related}
The explosion of fake news has emerged as a significant concern in the digital age, impacting public belief and individual behaviors. We explore various dimensions of fake news and provide a review by examining existing studies. The review focuses on traditional machine learning (ML), neural networks (NN), transformer-based neural models, and ensemble approaches. It identifies the scope and implications of fake news, as well as the gaps in the current state of the art that require further investigation. 

Traditional ML models such as Logistic Regression, Gradient Boosting, Support Vector Machines (SVM), Decision Tree, Random Forest, and Multinomial Naive Bayes, used for detecting misinformation in social media, face limitations in feature engineering, scalability, handling imbalanced data, contextual understanding, and adaptability to evolving patterns \cite{shu2017fake, kolesnikovadetecting, schutz2021automatic, yigezu2023habesha}. A study demonstrated the efficiency of ML approaches in discriminating between fake and genuine news with high accuracy by highlighting the importance of employing ML in combating the spread of fake news \cite{gupta2013faking}. Such studies have used linguistic and content-based features \cite{shu2017fake}. SVM, Naïve Bayes, and passive-aggressive classifiers are employed to detect fake news using the Term Frequency-Inverse Document Frequency (TF-IDF) as a feature extraction strategy and achieved an accuracy of 95.5\% \cite{shaikh2020fake}. SVM, Decision Trees, Naive Bayes, and Logistic Regression classifiers are employed to distinguish fake news from true news \cite{pandey2022fake}. Neural network models are better at extracting complex features compared to traditional ML models \cite{yigezu2023habeshab, janiesch2021machine, bade2021natural}. Several NN models, including convolutional neural networks (CNNs), recurrent neural networks (RNNs),  autoencoders, and generative adversarial neural networks(GAN),  have been used in many recent studies \cite{janiesch2021machine, yigezu2024habesha}. In particular, a CNN model is employed to detect fake news based on the discriminatory characteristics of its contents. Compared with other baseline models, it performs an accuracy of 98.36\% \cite{kaliyar2020fndnet}. Another study used an LSTMs NN with GloVe word embeddings to identify false information on social media. The authors enhanced their approach with \texttt{N-grams and tokenization }for feature extraction and demonstrated high performance, with an accuracy of 99.88\%.

Transformer-based neural models truly excel in capturing context for NLP tasks, an achievement that remains a challenge for basic NN and traditional ML models \cite{mersha2024semantic}. De et al. (2021) proposed an efficient model based on multilingual \texttt{BERT} for domain-agnostic multilingual fake news classification, which outperformed state-of-the-art models and achieved high accuracy across various domain-specific languages \cite{de2021transformer}. Schütz et al. (2021) developed an automatic binary classification method for identifying fake news using pre-trained transformer-based models like \texttt{XLNet, BERT, RoBERTa, DistilBERT, and ALBERT}. Their experiments demonstrated that transformers show promise in detecting fake news even with limited data \cite{schutz2021automatic, yigezuodio}.

Neural networks, including transformer-based models, are black-boxes in nature, and significant challenges arise, in regard to interpretability, accountability, transparency, and trust in the models' decision-making process and outputs \cite{arrieta2020explainable}. XAI is specifically designed to address these challenges. It provides clear insights into the model’s prediction and enhances the understanding and trust of the model users \cite{madhav2022explainable,  arrieta2020explainable}. Various XAI techniques can be employed to explain the model’s predictions for fake news detection. LIME and Anchor have been used to explain BERT model predictions in fake news detection by highlighting influential features \cite{szczepanski2021new}. Hashmi et al. (2024)  applied LIME to understand predictions from \texttt{CNN, LSTM, BERT, XLNet, and RoBERTa} models on WELFake, FakeNewsNet, and Fake NewsPrediction datasets \cite{hashmi2024advancing}. Dua et al. (2023) used \texttt{LIME and SHAP} to reveal key features in the predictions of \texttt{LSTM and BERT} models \cite{dua2023flash}.

Most approaches discussed in our review have used only the features of content text for fake news detection. The use of content text features has several limitations, such as failing to consider the broader context, struggling with language diversity, and ambiguity \cite{saquete2020fighting, mersha2024explainable}. Our study integrates social context with textual features to overcome these challenges. This combined approach improves the accuracy and robustness of fake news detection. We use Amharic language datasets specifically to address the language diversity challenge, ensuring that our model can effectively analyze and detect fake news across different languages.

\section{Data set and annotation} \label{data}

\begin{figure*}[h!]
\centering
\includegraphics[width=15 cm]{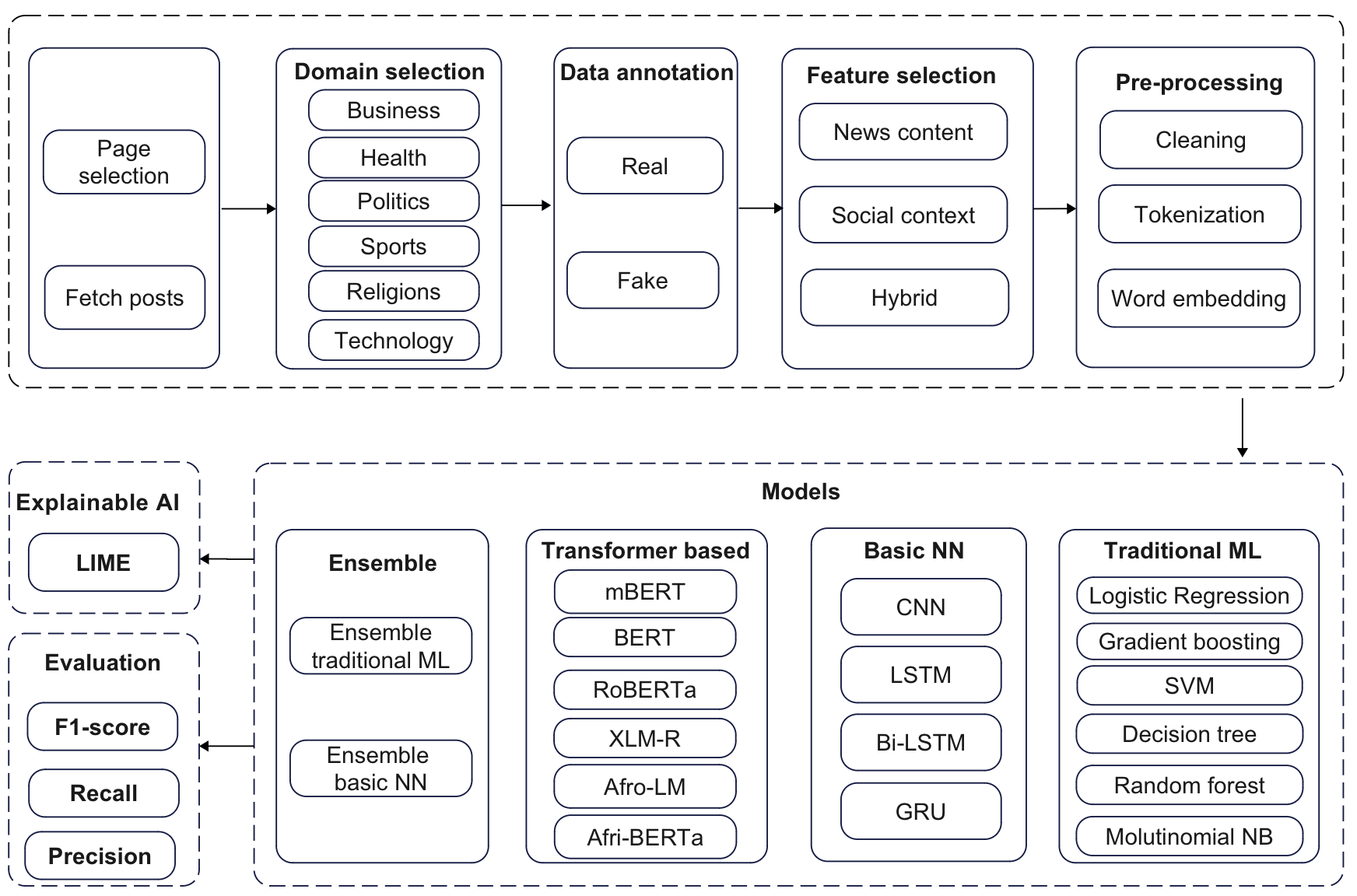}
\caption{Overview of the main phases of our methodology approach.
\label{fig-meth}}
\end{figure*}

Fig. \ref{fig-meth} presents cutting-edge approaches to combating fake news in under-resourced languages using explainable AI.

\begin{table}[h]
\caption{Our dataset domain statistics \label{domain}}
\begin{tabular*}{\hsize}{@{\extracolsep{\fill}}lccc@{}}
\toprule
\textbf{Domains} & \textbf{Real} & \textbf{Fake} & \textbf{Total} \\
\midrule
Business    & 932  & 1180 & 2112 \\
Health      & 763  & 204  & 967  \\
Politics    & 2000 & 2003 & 4003 \\
Sport       & 580  & 705  & 1285 \\
Religion    & 1455 & 1505 & 2960 \\
Technology  & 270  & 403  & 673  \\
\textbf{Total} & \textbf{6000} & \textbf{6000} & \textbf{12000} \\
\bottomrule
\label{table-1}
\end{tabular*}
\end{table}
Our study sourced data from various domains, including business, politics, sports, health, religion, and technology. The average length of the text is 128 characters. Table \ref{table-1} presents statistics on each topic within our corpus. We employed pre-processing techniques to improve data quality. Sensitive attributes were anonymized to protect privacy. We prepared data annotation guidelines to label the data effectively. To ensure label agreement, three annotators annotated the dataset based on these guidelines. To confirm the text's authenticity, we evaluated it by checking the source and author credentials.

\section{Models} \label{model}
We investigate various approaches to achieve better performance with hybrid features. We use different conventional algorithms, such as Logistic Regression, Gradient Boosting, SVM, Decision Tree, Random Forest and Multinomial NB that can play a crucial role in identifying false information. We use basic NN such as CNN, LSTM, BiLSTM and GRU and exploit their complementary strengths in capturing different aspects of textual information and enhancing the overall performance of fake news detection. We hypothesize that transfer learning approaches can improve the performance of the model by leveraging pre-trained language models that have already learned useful features from large datasets. These methods can reduce the amount of data and computational resources required for training and improve the generalization of the model. We consider \texttt{BERT} \cite{devlin2018bert} and \texttt{RoBERTa-base} \cite{liu2019roberta} as the baseline models for monolingual. \texttt{XLM-R \cite{conneau2019unsupervised}, Afro-LM} \cite{dossou2022afrolm}  and \texttt{Afri-BERTa} \cite{ogueji2021small} that are multilingual and cater to African languages. For exploring the impact of domain adaptive models, we experiment with \texttt{mBERT-Amharic} \footnote{https://huggingface.co/Davlan/bert-base-multilingual-cased-finetuned-amharic}. We perform ensemble learning by combining multiple traditional ML and basic NN methods to enhance predictive performance and robustness. We use XGBoost \cite{chen2016xgboost} as the meta classifier to make the final predictions.

We extract features from both social contexts and news content, aiming to capture the dynamic interactions within social networks. The dataset is classified into 80\% for training and 20\% for testing. Various metrics, including precision, recall, and the F1 score, serve as benchmarks for evaluating the performance of models.

\section{Experimental Results} \label{result}

In the tables given below, we present the experimental outcomes of various methods. The goal is to discern fake news within social media posts, and the results are presented separately each category of the employed approach.

\begin{table}[h]
\caption{Experimental results of traditional ML approaches on test data.\label{ml}}
\begin{tabular*}{\hsize}{@{\extracolsep{\fill}}lccc@{}}
\toprule
\textbf{Models} & \textbf{Precision} & \textbf{Recall} & \textbf{F1 score} \\
\midrule
Logistic Regression  & 0.93 & 0.92 & 0.92 \\
Gradient Boosting    & 0.88 & 0.87 & 0.86 \\
\textbf{SVM}         & \textbf{0.93} & \textbf{0.93} & \textbf{0.93} \\
Decision Tree        & 0.87 & 0.87 & 0.87 \\
Random Forest        & 0.91 & 0.90 & 0.90 \\
Multinomial NB       & 0.93 & 0.93 & 0.93 \\
\bottomrule
\end{tabular*}
\end{table}

Ultimately, a comparative analysis is conducted to determine the most effective models or techniques as per the findings of this study.
As indicated in Table \ref{ml}, various experiments were conducted using ML methodologies. These experiments involved the analysis of texts with a multitude of features. Notably, the SVM technique surpassed others, achieving a macro F1-score of 0.93. Logistic Regression and NB exhibit commendable performance, closely trailing behind SVM with F1 scores of 0.92. The remaining models have performed as follows: Random Forest achieve a score of 0.9, Decision Tree scored 0.87, and Gradient Boosting attained 0.86.

The presented Table \ref{dl} depicts the experiments carried out utilizing various basic NN methods. It is evident from the results that the GRU model outperforms other models, achieving a macro F1-score of 0.89. Following closely behind, the Bi-LSTM displays strong performance with an F1-score of 0.88.


\begin{table}[h]
\caption{Experimental results of NN approaches on test dataset.\label{dl}}
\begin{tabular*}{\hsize}{@{\extracolsep{\fill}}lccc@{}}
\toprule
\textbf{Models} & \textbf{Precision} & \textbf{Recall} & \textbf{F1 score} \\
\midrule
CNN         & 0.88 & 0.88 & 0.87 \\
Bi-LSTM     & 0.88 & 0.88 & 0.88 \\
LSTM        & 0.87 & 0.87 & 0.87 \\
\textbf{GRU} & \textbf{0.89} & \textbf{0.89} & \textbf{0.89} \\
\bottomrule
\end{tabular*}
\end{table}

The CNN and LSTM models show comparable results, each obtaining an F1-score of 0.87. Among the basic NN approaches, GRU effectively discerns fake news from social media posts for several reasons.


\begin{table}[h]
\caption{Experimental results of transfer learning approaches on the test dataset.\label{tl}}
\begin{tabular*}{\hsize}{@{\extracolsep{\fill}}lccc@{}}
\toprule
\textbf{Models} & \textbf{Precision} & \textbf{Recall} & \textbf{F1 score} \\
\midrule
\texttt{\textbf{Fine-tuned mBERT}} & \textbf{0.95} & \textbf{0.93} & \textbf{0.94} \\
\texttt{BERT}                      & 0.92         & 0.91          & 0.90          \\
\texttt{RoBERTa}                   & 0.90         & 0.90          & 0.90          \\
\texttt{XLM-R}                     & 0.78         & 0.69          & 0.67          \\
\texttt{Afro-LM}                   & 0.83         & 0.83          & 0.83          \\
\texttt{Afri-BERTa}                & 0.89         & 0.89          & 0.89          \\
\bottomrule
\end{tabular*}
\end{table}

We conduct various experiments using pre-trained language models. As indicated in Table \ref{tl}, we investigate the performance of LM \texttt{BERT, RoBERTa-base, and fine-tuned mBERT Amharic}, multilingual pre-trained models that are trained with Afro-centric languages, including Amharic.

Upon evaluation, we achieve better results with monolingual LM, specifically the fine-tuned \texttt{mBERT Amharic, BERT, and RoBERTa-base} models, achieve macro F1 scores of 0.94, 0.9, and 0.9, respectively. Furthermore, \texttt{Afri-BERTa} outperformed \texttt{Afro-LM and XLM-R}, achieving a macro F1-score of 0.89, while \texttt{Afro-LM and XLM-R} yielded successive performances with macro F1 scores of 0.83 and 0.67, respectively.

From these results, we observe that when pre-trained language models are trained on a smaller number of languages, they perform better than models trained on a larger number of languages. The fine-tuned \texttt{mBERT Amharic} model outperformed other models. Next to the fine-tuned \texttt{Amharic mBERT model, BERT and RoBERTa} as monolingual language models, outperformed other models. This indicates that monolingual models, being exclusively trained on data from a single language, are able to grasp intricate details and nuances specific to that language, leading to improved performance in tasks requiring precise comprehension and generation of text in that language. Specifically, a monolingual model that is fine-tuned in the targeted language performed well and accurately compared to other monolingual models.

\texttt{XLM-R} performed poorly in this experiment, likely due to being trained on a large dataset encompassing 100 languages. This suggests that multilingual models, which must accommodate multiple languages within a single architecture, may experience interference or dilution of language-specific features. In contrast, monolingual models maintain a focus solely on one language, thereby avoiding potential conflicts or compromises in representation.


\begin{table}[h]
\caption{Experimental results of ensemble learning approaches on the test dataset.\label{el}}
\begin{tabular*}{\hsize}{@{\extracolsep{\fill}}lccc@{}}
\toprule
\textbf{Models} & \textbf{Precision} & \textbf{Recall} & \textbf{F1 score} \\
\midrule
Ensemble ML  & 0.99 & 0.99 & 0.99 \\
Ensemble NN  & 0.98 & 0.98 & 0.98 \\
\bottomrule
\end{tabular*}
\end{table}

To enhance the accuracy of identifying fake news, we employ ensemble techniques using the stacking method. Initially, we use various traditional machine learning and neural network techniques to train models on an initial dataset and consolidate them by merging the top three probabilities into a single matrix. We used their predictions to create a new dataset for training a meta-model, which then produces a final prediction through weighted averaging. This achieves an impressive F1 score of 0.99 and 0.98, respectively. These approaches surpass standalone traditional ML and basic NN in terms of performance. The results in Table \ref{el} emphasize the efficacy of ensemble learning in enhancing predicts accuracy across diverse traditional machine learning tasks.

\section{Interpretability modeling} \label{xai}
Interpretability modeling is crucial in identifying fake news because it clarifies the complex processes of models. Understanding the reasoning behind classifications helps identify biases, assess model reliability, and refine approaches. This transparency builds trust and accountability, essential for combating misinformation. Local Interpretable Model-agnostic Explanations (LIME) are valuable as they explain specific predictions, highlighting factors that lead to classifying news articles as fake \cite{lundberg2017unified}. We chose the LIME explainer for its human-understandable explanations and visualization capability for low-resource language. This facilitates the assessment of the model's decision-making process by non-experts, such as journalists or policymakers, who are key stakeholders in the fight against fake news. LIME is also effective for text-based LLMs and produces consistent and stable explanations \cite{atanasova2024diagnostic}.

\begin{figure*}[h!]
\centering
\begin{subfigure}[b]{0.5\textwidth}
    \centering
    \includegraphics[width=\textwidth]{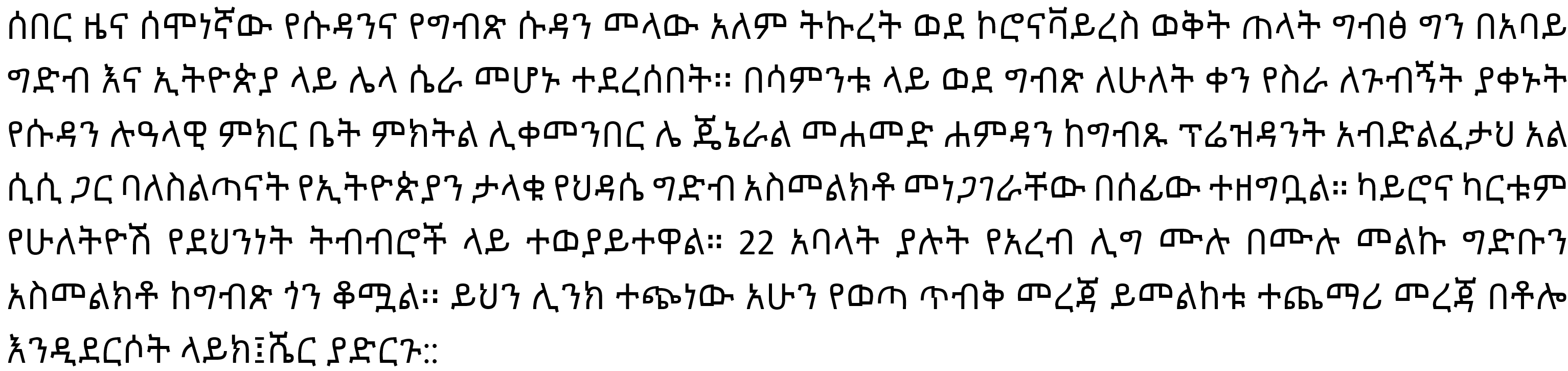}
    \caption{Sample Amharic input text used to demonstrate the decision-making processes of various models with LIME explainer.}
    \label{fig-inputText1}
\end{subfigure}
\hfill
\begin{subfigure}[b]{0.45\textwidth}
    \centering
    \includegraphics[width=\textwidth]{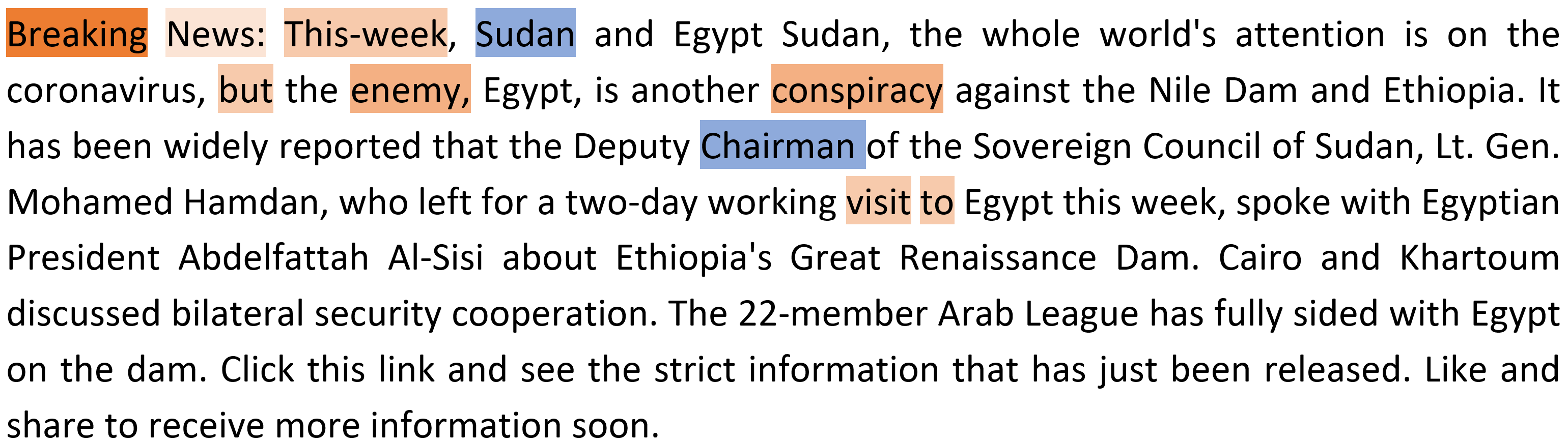}
    \caption{ English translation of the Amharic input text with highlighted words indicating equivalent LIME explanations for the AfriBERTa model.}
    \label{fig-inputText2}
\end{subfigure}
\caption{a) Amharic input text; b) English translation and  LIME explanation.}
\label{fig-combined}
\end{figure*}

\begin{figure*}[h!]
\centering
\includegraphics[width=13 cm]{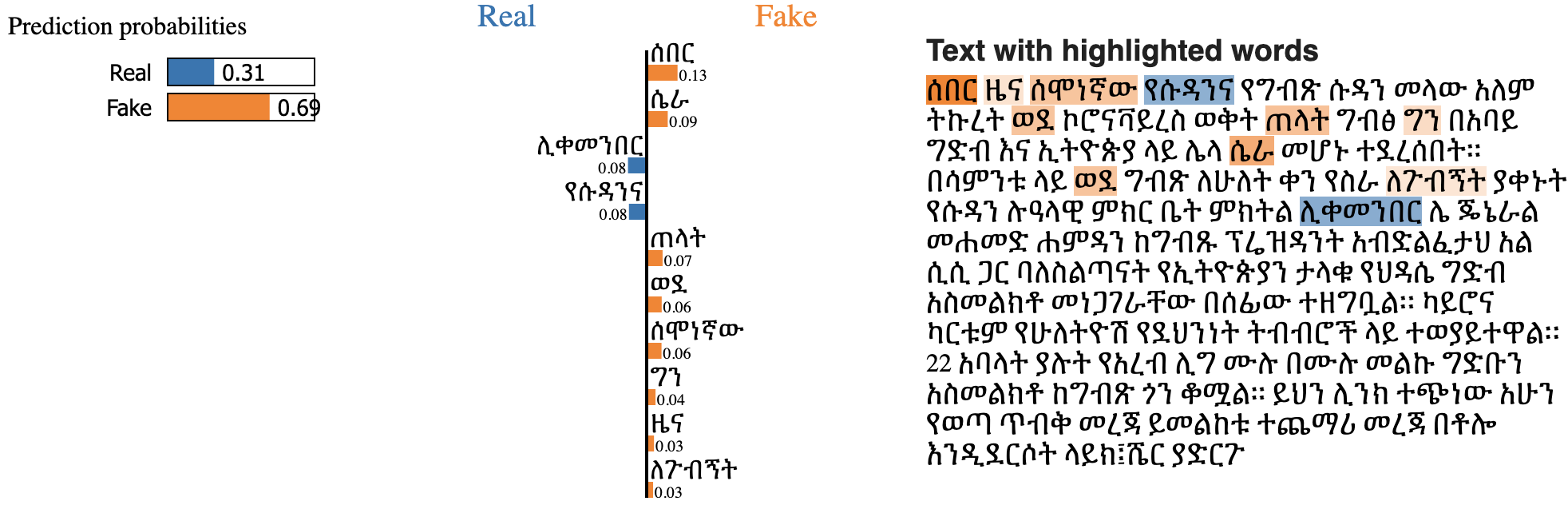}

\caption{Sample LIME explanations demonstrating feature importance in \texttt{Afri-Berta} model}
\label{fig-lstm}
\end{figure*}

Through our analysis, we aimed to understand the models' predictions and identify biases. Using LIME, we gained insights into model workings and improved interpretability, fostering trust and understanding. We observed that different models identified distinct key features influencing their predictions, highlighting their varied decision-making processes. By comparing basic neural and advanced transformer-based models on a common input text, Fig.\ref{fig-combined}, we found that all predicted it as 'fake' news, but each used different significant features. Table \textcolor{blue}{6-8} shows that while models may reach the same conclusion, their reasoning and feature importance vary due to their internal structures. Fig. \ref{fig-lstm} illustrates how the LIME technique identifies important features considering the decision made by LSTM algorithm.


\begin{figure*}[h!]
\centering
\includegraphics[width=12 cm]{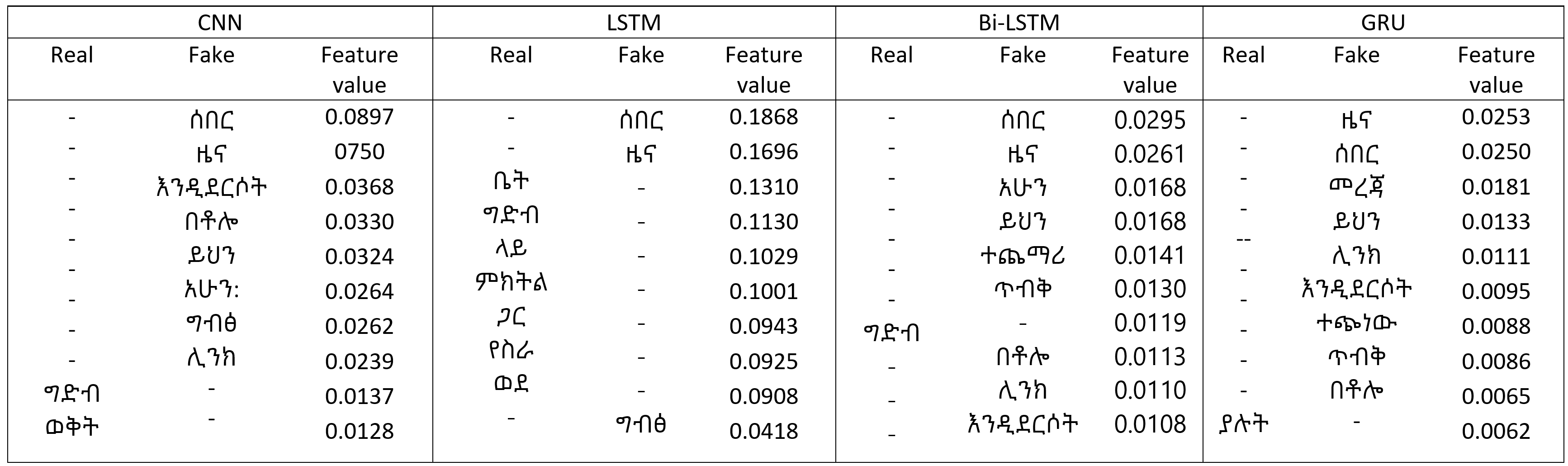}

\text{Table 6: Summary of the key feature words derived from LIME for basic neural network models.}
\label{fig-NN}
\end{figure*}

\begin{figure*}[h!]
\centering
\includegraphics[width=12 cm]{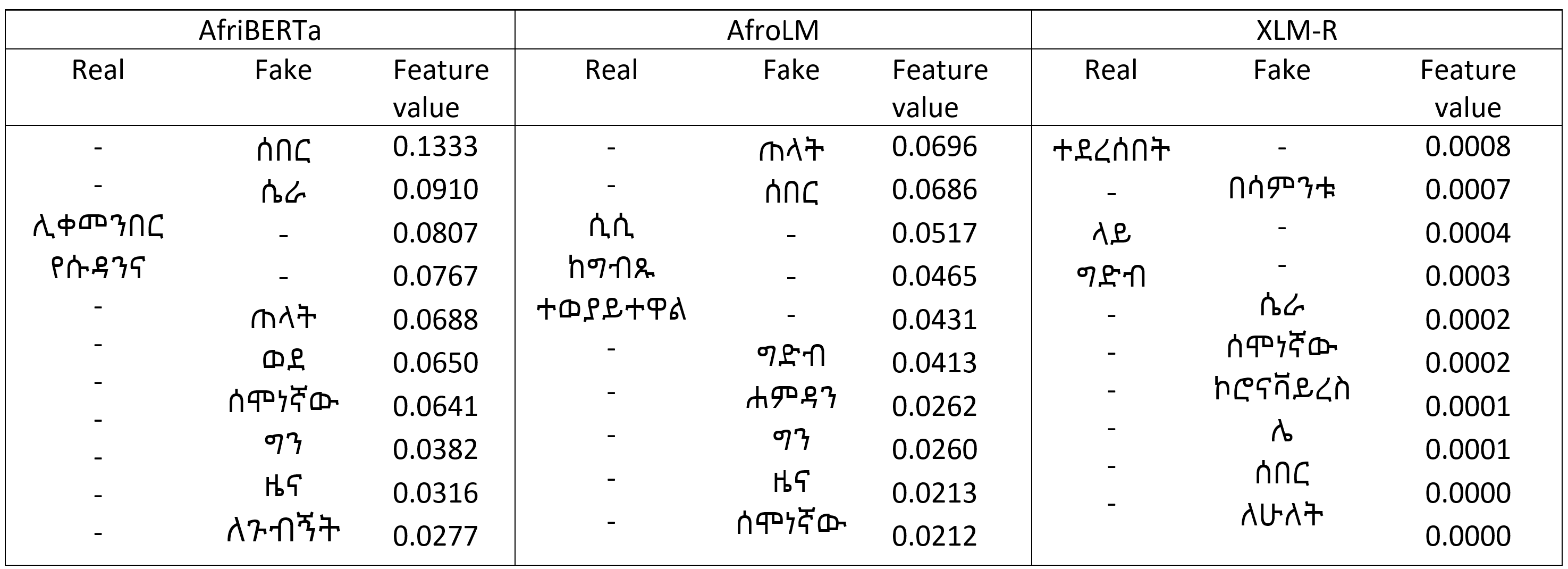}\\
\small\text{Table 7: Summary of the key feature words derived from LIME for transformer-based models}
\label{fig-trans}
\end{figure*} 

\begin{figure*}[h!]
\centering
\includegraphics[width=12 cm]{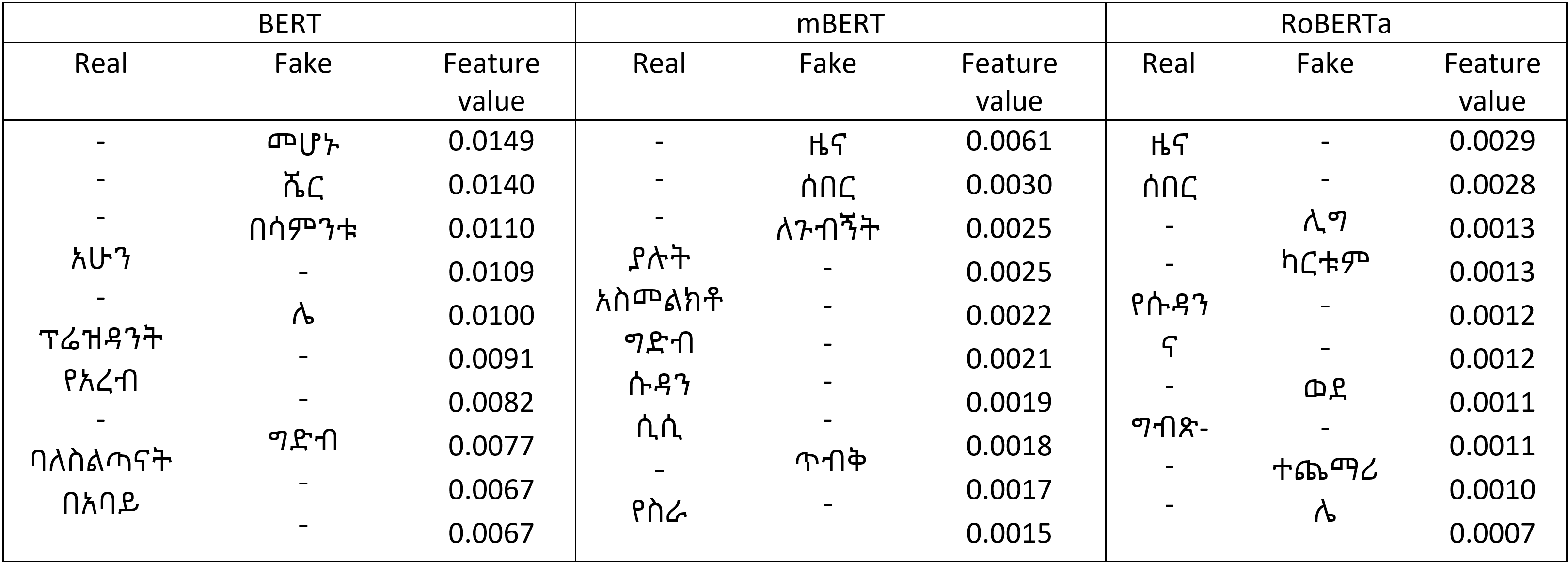}\\
\text{Table 8: Summary of the key feature words derived from LIME for transformer-based models.}
\label{fig-tl}
\end{figure*} 

\section{Conclusion} \label{conclusion}

In response to the widespread dissemination of fake news, particularly in Ethiopia, where content regulation is lacking, there is a critical need for effective fake news detection mechanisms. Despite this urgency, existing efforts encounter challenges like the absence of annotated datasets and insufficient attention to fake news features.

To address these issues, we have developed a comprehensive Facebook-sourced corpus covering various domains and integrated diverse fake news features to enhance detection capabilities. We then built a sophisticated fake news detection model using traditional machine learning, basic neural networks, ensemble learning, and transfer learning. The fine-tuned Amharic mBERT model outperformed existing methods, highlighting the effectiveness of language-specific models.

We extensively evaluated neural network and transformer-based architectures using the LIME technique to gauge explainability. These efforts contribute to a robust and explainable framework for combating fake news dissemination on social media platforms. This method facilitates this understanding by providing insights into how algorithms arrive at their decisions.

\section*{Acknowledgments}
The work was done with partial support from the Mexican Government through the grant A1-S-47854 of CONACYT, Mexico,
grants 20241816, 20241819,  and 20240951 of the Secretaría de Investigación y Posgrado of the Instituto Politécnico Nacional, Mexico. The authors thank the CONACYT for the computing resources brought to them through the Plataforma de Aprendizaje Profundo para Tecnologías del Lenguaje of the Laboratorio de Supercómputo of the INAOE, Mexico and acknowledge the support of Microsoft through the Microsoft Latin America PhD Award.

\bibliographystyle{elsarticle-num} 
 \bibliography{cas-refs}

\end{document}